\documentclass[10pt,journal,compsoc, hidelinks]{IEEEtran}

\usepackage{graphicx}
\usepackage{amsmath}
\usepackage{amsfonts}
\usepackage{caption}
\usepackage{subcaption}
\usepackage{combelow}
\usepackage{todonotes}
\usepackage[symbol*]{footmisc}
\usepackage{float}

\newcommand{\paraV}{\vspace{1em}}

\begin{document}



\title{DeXpression: Deep Convolutional Neural Network for Expression Recognition}



\author{

Peter Burkert\IEEEauthorrefmark{1}\IEEEauthorrefmark{3},
Felix Trier\IEEEauthorrefmark{1}\IEEEauthorrefmark{3},
Muhammad Zeshan Afzal\IEEEauthorrefmark{2}\IEEEauthorrefmark{3},

Andreas Dengel\IEEEauthorrefmark{2}\IEEEauthorrefmark{3} and 
Marcus Liwicki\IEEEauthorrefmark{3}

\IEEEauthorblockA{\IEEEauthorrefmark{2}German Research Center for Artificial Intelligence (DFKI), Kaiserslautern, Germany}

\IEEEauthorblockA{\IEEEauthorrefmark{3}University of Kaiserslautern, Gottlieb-Daimler-Str., Kaiserslautern 67663, Germany\\p\_burkert11@cs.uni-kl.de, f\_trier10@cs.uni-kl.de,  afzal@iupr.com, andreas.dengel@dfki.de, liwicki@dfki.uni-kl.de}

}

\maketitle

\begin{abstract}

We propose a convolutional neural network (CNN) architecture for facial expression recognition. The proposed architecture is independent of any hand-crafted feature extraction and performs better than the earlier proposed convolutional neural network based approaches. We visualize the automatically extracted features which have been learned by the network in order to provide a better understanding. The standard datasets, i.e. Extended Cohn-Kanade (CKP) and MMI Facial Expression Databse are used for the quantitative evaluation. On the CKP set the current state of the art approach, using CNNs, achieves an accuracy of 99.2\%. For the MMI dataset, currently the best accuracy for emotion recognition is 93.33\%. The proposed architecture achieves $99.6$\% for CKP and $98.63$\% for MMI, therefore performing better than the state of the art using CNNs. Automatic facial expression recognition has a broad spectrum of applications such as human-computer interaction and safety systems. This is due to the fact that non-verbal cues are important forms of communication and play a pivotal role in interpersonal communication. The performance of the proposed architecture endorses the efficacy and reliable usage of the proposed work for real world applications.

\end{abstract}

\section{Introduction}

Humans use different forms of communications such as speech, hand gestures and emotions. Being able to understand one's emotions and the encoded feelings is an important factor for an appropriate and correct understanding.

With the ongoing research in the field of robotics, especially in the field of humanoid robots, it becomes interesting to integrate these capabilities into machines allowing for a more diverse and natural way of communication. One example is the Software called EmotiChat~\cite{Anderson06areal-time}. This is a chat application with emotion recognition. The user is monitored and whenever an emotion is detected (smile, etc.), an emoticon is inserted into the chat window. Besides Human Computer Interaction other fields like surveillance or driver safety could also profit from it. Being able to detect the mood of the driver could help to detect the level of attention, so that automatic systems can adapt.\\
\let\thefootnote\relax\footnote{*F. Trier and P. Burkert contributed equally to this work.}

Many methods rely on extraction of the facial region. This can be realized through manual inference~\cite{4032815} or an automatic detection approach~\cite{Anderson06areal-time}.
Methods often involve the Facial Action Coding System (FACS) which describes the facial expression using Action Units (AU). An Action Unit is a facial action like "raising the Inner Brow". Multiple activations of AUs describe the facial expression~\cite{kumar2009face}. Being able to correctly detect AUs is a helpful step, since it allows making a statement about the activation level of the corresponding emotion. \\
Handcrafted facial landmarks can be used such as done by Kotsia et al.~\cite{4032815}. Detecting such landmarks can be hard, as the distance between them differs depending on the person~\cite{6998925}. Not only AUs can be used to detect emotions, but also texture. When a face shows an emotion the structure changes and different filters can be applied to detect this~\cite{6998925}.\\

\begin{figure}
   \centering
        \includegraphics[width=\columnwidth]{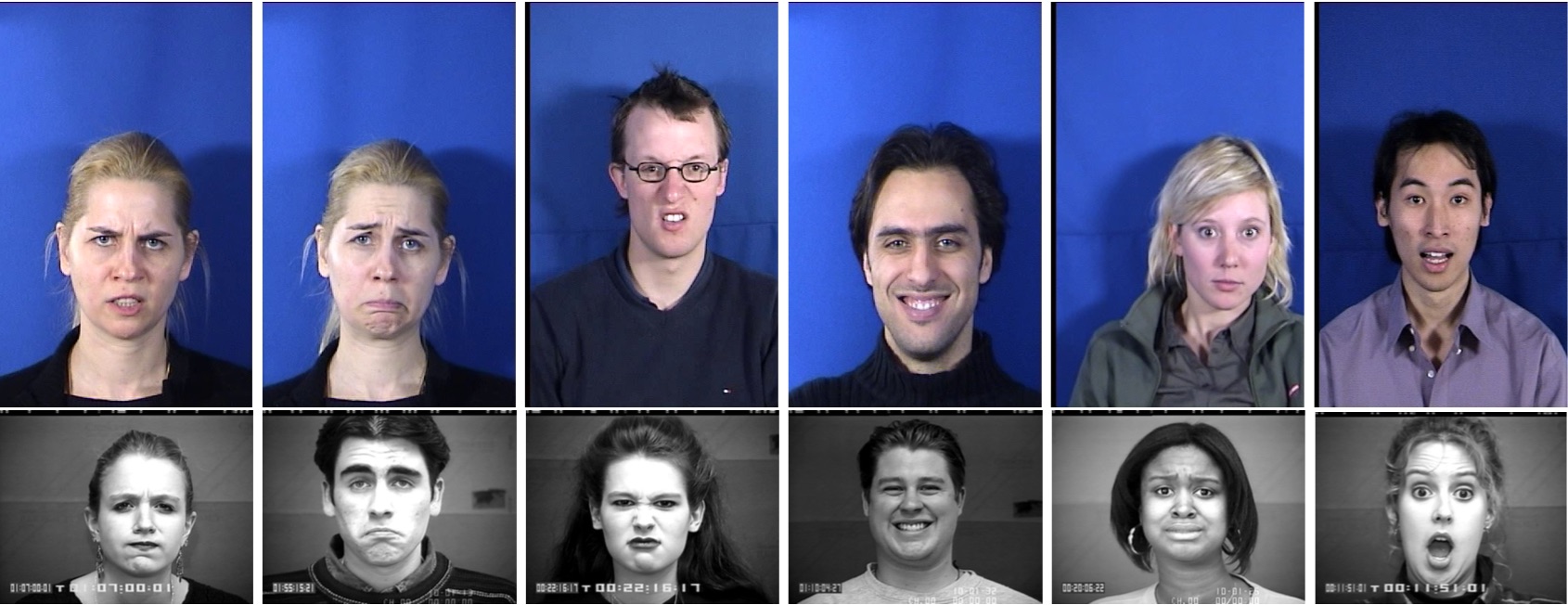}
   \caption{Example images from the MMI (top) and CKP (bottom). The emotions from left to right are: \textit{Anger}, \textit{Sadness}, \textit{Disgust}, \textit{Happiness}, \textit{Fear}, \textit{Surprise}. The emotion \textit{Contempt} of the CKP set is not displayed.}\label{fig:example_images}
\end{figure}

The presented approach uses Artificial Neural Networks (ANN). ANNs differ, as they are trained on the data with less need for manual interference. 
Convolutional Neural Networks are a special kind of ANN and have been shown to work well as feature extractor when using images as input~\cite{donahue2013decaf} and are real-time capable. This allows for the usage of the raw input images without any pre- or postprocessing.\\
GoogleNet~\cite{DBLP:journals/corr/SzegedyLJSRAEVR14} is a deep neural network architecture that relies on CNNs. It has been introduced during the Image Net Large Scale Visual Recognition Challenge(ILSVRC) 2014. This challenge analyses the quality of different image classification approaches submitted by different groups. The images are separated into 1000 different classes organized by the WordNet hierarchy. In the challenge "object detection with additional training data" GoogleNet has achieved about 44\% precision~\cite{LSVRC-results}. These results have demonstrated the potential which lies in this kind of architecture. Therefore it has been used as inspiration for the proposed architecture.\\
The proposed network has been evaluated on the Extended Cohn-Kanade Dataset (Section~\ref{sec:ckp}) and on the MMI Dataset (Section~\ref{sec:mmi}). Typical pictures of persons showing emotions can be seen in Fig.~\ref{fig:example_images}.
The emotion \textit{Contempt} of the CKP set is not shown as no subject with consent for publication and an annotated emotion is part of the dataset. Results of experiments on these datasets demonstrate the success of using a deep layered neural network structure. With a 10-fold cross-validation a recognition accuracy of 99.6\% has been achieved. \\

The paper is arranged as follows: After this introduction, Related Work (Section~\ref{sec:related}) is presented which focuses on Emotion/Expression recognition and the various approaches scientists have taken. Next is Section~\ref{sec:background}, Background, which focuses on the main components of the architecture proposed in this article. Section~\ref{sec:datasets} contains a summary of the used Datasets. In Section~\ref{sec:architecture} the architecture is presented. This is followed by the experiments and its results (Section~\ref{sec:experiments}) . Finally, Section~\ref{sec:conclusion} summarizes the article and concludes the article.
\section{Related Work}
\label{sec:related}
A detailed overview for expression recognition was given by C\u{a}leanu~\cite{caleanu2013face} and Bettadapura~\cite{bettadapura2012face}. In this Section mainly work which similar to the proposed method is presented as well as few selected articles which give a broad overview over the different methodologies.\\

Recently Szegedy et al.\cite{DBLP:journals/corr/SzegedyLJSRAEVR14} have proposed an architecture called GoogLeNet. This is a 27 layer deep network, mostly composed of CNNs. The network is trained using stochastic gradient descent. In the ILSVRC 2014 Classification Challenge this network achieved a top-5 error rate of 6.67\% winning the first place. \\
Using the the Extended Cohn-Kanade Dataset (Section~\ref{sec:ckp}), Happy and Routray~\cite{6998925} classify between six basic emotions. Given an input image, their solution localizes the face region. From this region, facial patches e.g. the eyes or lips are detected and points of interest are marked. From the patches which have the most variance between two images, features are extracted. The dimensionality of the features is reduced and then given to a Support Vector Machine (SVM). To evaluate the method, a 10-fold cross-validation is applied. 
The average accuracy is 94.09\%.\\
Video based emotion recognition has been proposed by Byeon and Kwak~\cite{byeonfacial}. They have developed a three dimensional CNN which uses groups of 5 consecutive frames as input. A database containing 10 persons has been used to achieve an accuracy of 95\%.\\
Song et al.~\cite{song2014deep} have used a deep convolutional neural network for learning facial expressions. The created network consists of five layers with a total of 65k neurons. Convolutional, pooling, local filter layers and one fully connected layer are used to achieve an accuracy of 99.2\% on the CKP set. To avoid overfitting the dropout method was used.\\
Luecy et al.~\cite{5543262} have created the Extended Cohn-Kanade dataset. This dataset contains emotion annotations as well as Action Unit annotations. In regards to classification, they also have evaluated the datasets using Active Appearance Models (AAMs) in combination with SVMs. To find the position and track the face over different images, they have employed AAM which generates a Mesh out of the face. From this mesh they have extracted two feature vectors. First, the normalized vertices with respect to rotation, translation, and scale. Second a gray-scale image from the mesh data, and the input images has been extracted. They have chosen a cross-validation strategy, where one subject is left out in the training process, achieving an accuracy of over 80\%.\\
Anderson et al.~\cite{Anderson06areal-time} have developed a face expression system, which is capable of recognizing the six basic emotions. Their system is built upon three components. The first one is a face tracker (derivative of ratio template) to detect the location of the face. The second component is an optical flow algorithm to track the motion within the face. The last component is the recognition engine itself. It is based upon Support Vector Machines and Multilayer Perceptrons. This approach has been implemented in EmotiChat. They achieve a recognition accuracy of 81.82\%.\\
Kotsia and Pitas~\cite{4032815} detect emotions by mapping a Candide grid, a face mask with a low number of polygons, onto a person's face. The grid is initially placed randomly on the image, then it has to be manually placed on the persons face. 
Throughout the emotion, the grid is tracked using a Kanade–Lucas–Tomasi tracker. The geometric displacement information provided by the grid is used as feature vector for multiclass SVMs. The emotions are anger, disgust, fear, happiness, sadness, and surprise. They evaluate the model on the Cohn-Kanade dataset and an accuracy of 99.7\% has been achieved.\\
Shan et al.~\cite{Shan2009803} have created an emotion recognition system based on Local Binary Patterns (LBP). The LBPs are calculated over the facial region. From the extracted LBPs a feature vector is derived. The features depend on the position and size of the sub-regions over witch the LBP is calculated. AdaBoost is used to find the sub-regions of the images which contain the most discriminative information. Different classification algorithms have been evaluated of which an SVM with Boosted-LBP features performs the best with a recognition accuracy of 95.1\% on the CKP set.\\
In 2013 Zafar et al.~\cite{6743520} proposed an emotion recognition system using Robust Normalized Cross Correlation (NCC). The used NCC is the "Correlation as a Rescaled Variance of the Difference between Standardized Scores". Outlier pixels which influence the template matching too strong or too weak are excluded and not considered. This approach has been evaluated on different databases including AR FaceDB (85\% Recognition Accuracy) and the Extended Cohn Kanade Database (100\% Recognition Accuracy).\\

\section{Convolutional Neural Networks}
\label{sec:background}

\paraV
\paragraph{\textit{Convolutional Layer}}
Convolutional Layers perform a convolution over the input. Let $f_k$ be the filter with a kernel size $n\times m$ applied to the input $x$. $n \times m$ is the number of input connections each CNN neuron has. The resulting output of the layer calculates as follows: 

\begin{center}
\begin{minipage}{0.8\columnwidth}
\begin{equation}
C(x_{u,v}) = \sum_{i=-\frac{n}{2}}^{\frac{n}{2}}\sum_{j=-\frac{m}{2}}^{\frac{m}{2}}f_k(i,j) x_{u-i,v-j}
\end{equation}
\end{minipage}
\end{center}

To calculate a more rich and diverse representation of the input, multiple filters $f_k$ with $k \in \mathbb{N}$ can be applied on the input. The filters $f_k$ are realized by sharing weights of neighboring neurons. This has the positive effect that lesser weights have to be trained in contrast to standard Multilayer Perceptrons, since multiple weights are bound together.

\paraV
\paragraph{\textit{Max Pooling}}

Max Pooling reduces the input by applying the maximum function over the input $x_i$. Let $m$ be the size of the filter, then the output calculates as follows:

\begin{center}
\begin{minipage}{1\columnwidth}
\begin{equation}
M(x_i) = \max\{x_{i+k, i+l} \mid |k| \leq \frac{m}{2}, |l| \leq \frac{m}{2}\ k,l \in \mathbb{N}\}
\end{equation}
\end{minipage}
\end{center}

This layer features translational invariance with respect to the filter size.

\paraV
\paragraph{\textit{Rectified Linear Unit}}
A Rectified Linear Unit (ReLU) is a cell of a neural network which uses the following activation function to calculate its output given $x$:

\begin{center}
\begin{minipage}{0.44\columnwidth}
\begin{equation}
R(x) = max(0, x)
\end{equation}
\end{minipage}
\end{center}

Using these cells is more efficient than sigmoid and still forwards more information compared to binary units. When initializing the weights uniformly, half of the weights are negative. This helps creating a sparse feature representation. Another positive aspect is the relatively cheap computation. No exponential function has to be calculated. This function also prevents the vanishing gradient error, since the gradients are linear functions or zero but in no case non-linear functions~\cite{AISTATS2011_GlorotBB11}.

\paraV
\paragraph{\textit{Fully Connected Layer}}

The fully connected layer also known as Multilayer Perceptron connects all neurons of the prior layer to every neuron of its own layer. Let the input be $x$ with size $k$ and $l$ be the number of neurons in the fully connected layer. This results in a Matrix $W_{l \times k}$.

\begin{center}
\begin{minipage}{0.4\columnwidth}
\begin{equation}
F(x) = \sigma(W*x)
\end{equation}
\end{minipage}
\end{center}

$\sigma$ is the so called activation function. In our network $\sigma$ is the identity function. 

\paraV
\paragraph{\textit{Output Layer}}
The output layer is a one hot vector representing the class of the given input image. It therefore has the dimensionality of the number of classes. The resulting class for the output vector $x$ is:

\begin{center}
\begin{minipage}{0.65\columnwidth}
\begin{equation}
C(x) = \{i\ |\ \exists i \forall j \neq i : x_j \leq x_i\}
\end{equation}
\end{minipage}
\end{center}

\paraV
\paragraph{\textit{Softmax Layer}}
The error is propagated back over a Softmax layer. Let N be the dimension of the input vector, then Softmax calculates a mapping such that:
$S(x): \mathbb{R}^N \rightarrow [0,1]^N$

For each component $1 \leq j \leq N$, the output is calculated as follows:

\begin{center}
\begin{minipage}{0.5\columnwidth}
\begin{equation}
S(x)_j = \frac{e^{x_j}}{\sum_{i=1}^Ne^{x_i}}
\end{equation}
\end{minipage}
\end{center}

\section{Datasets}
\label{sec:datasets}
\newcommand{\mmiColWidth}{0.15\columnwidth}
\newcommand{\hspacing}{15mm}
\subsection{MMI Dataset}
\label{sec:mmi}
The MMI dataset has been introduced by Pantic et al.~\cite{Pantic2005wdffe} contains over 2900 videos and images of 75 persons.
The annotations contain action units and emotions. The database contains a web-interface with an integrated search to scan the database. The videos/images are colored. The people are of mixed age, different gender and have different ethnical background.
The emotions investigated are the six basic emotions: \textit{Anger}, \textit{Disgust}, \textit{Fear}, \textit{Happiness}, \textit{Sadness}, \textit{Surprise}.

\begin{figure}
\centering

    \begin{subfigure}[b]{\mmiColWidth}
      \includegraphics[width=\textwidth]{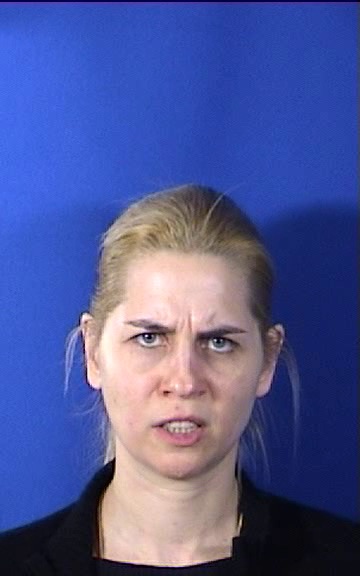}
    \end{subfigure}
    \begin{subfigure}[b]{\mmiColWidth}
      \includegraphics[width=\textwidth]{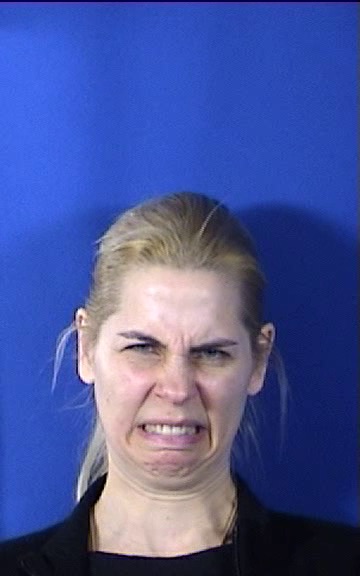}
    \end{subfigure}
    \begin{subfigure}[b]{\mmiColWidth}
      \includegraphics[width=\textwidth]{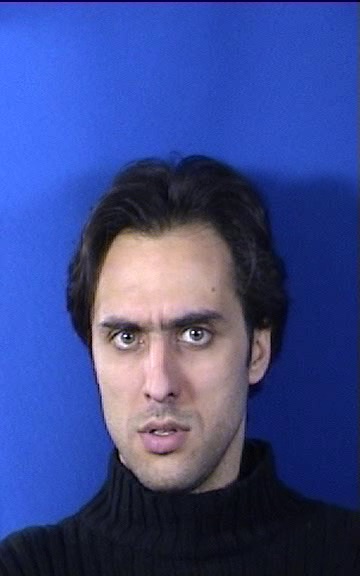}
    \end{subfigure}
    \begin{subfigure}[b]{\mmiColWidth}
      \includegraphics[width=\textwidth]{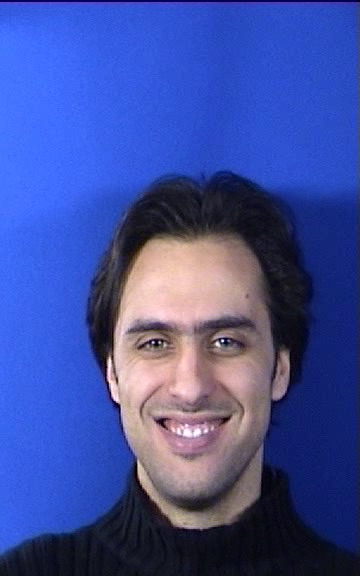}
    \end{subfigure}
    \begin{subfigure}[b]{\mmiColWidth}
      \includegraphics[width=\textwidth]{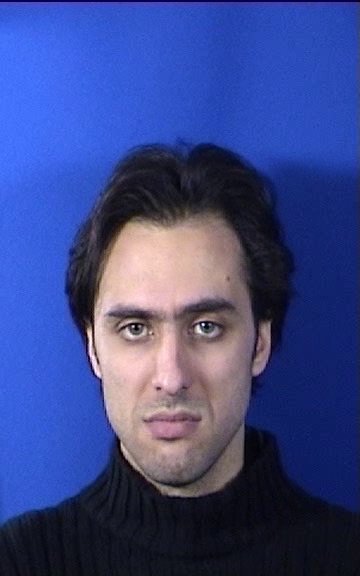}
    \end{subfigure}
    \begin{subfigure}[b]{\mmiColWidth}
      \includegraphics[width=\textwidth]{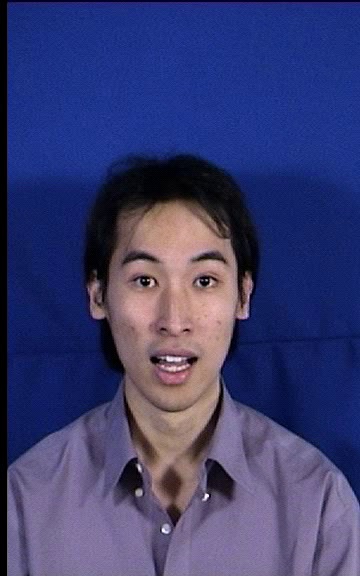}
    \end{subfigure}

    \begin{subfigure}[b]{\mmiColWidth}
      \includegraphics[width=\textwidth]{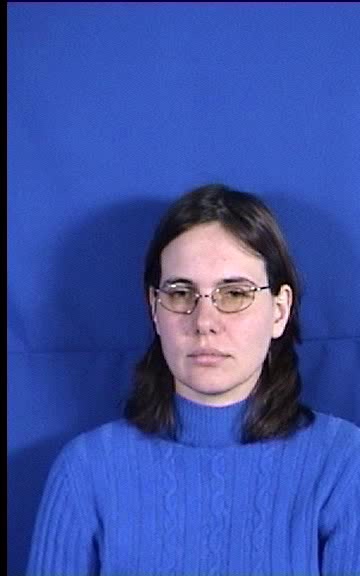}
      \caption*{Anger}
    \end{subfigure}
    \begin{subfigure}[b]{\mmiColWidth}
      \includegraphics[width=\textwidth]{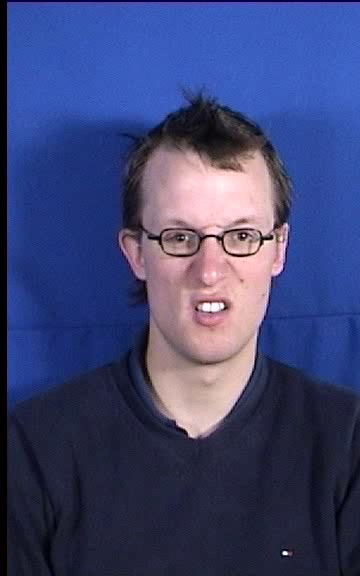}
      \caption*{Disgust}
    \end{subfigure}
    \begin{subfigure}[b]{\mmiColWidth}
      \includegraphics[width=\textwidth]{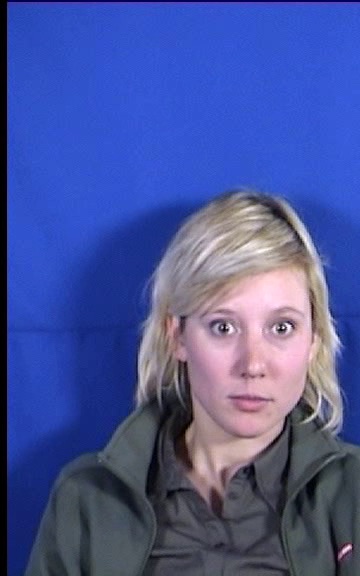}
      \caption*{Fear}
    \end{subfigure}
    \begin{subfigure}[b]{\mmiColWidth}
      \includegraphics[width=\textwidth]{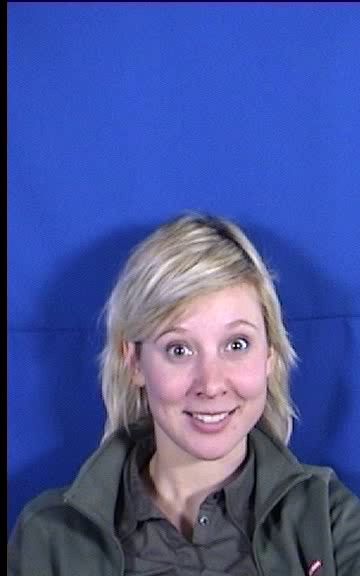}
      \caption*{Happy}
    \end{subfigure}
    \begin{subfigure}[b]{\mmiColWidth}
      \includegraphics[width=\textwidth]{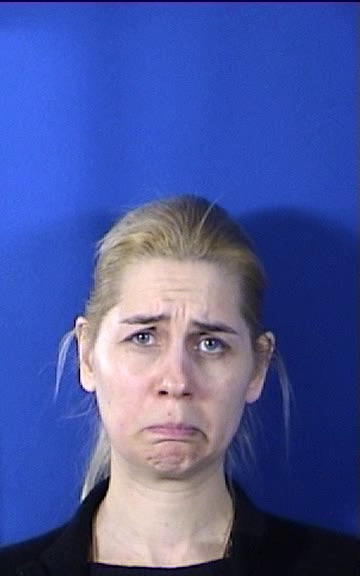}
      \caption*{Sadness}
    \end{subfigure}
    \begin{subfigure}[b]{\mmiColWidth}
      \includegraphics[width=\textwidth]{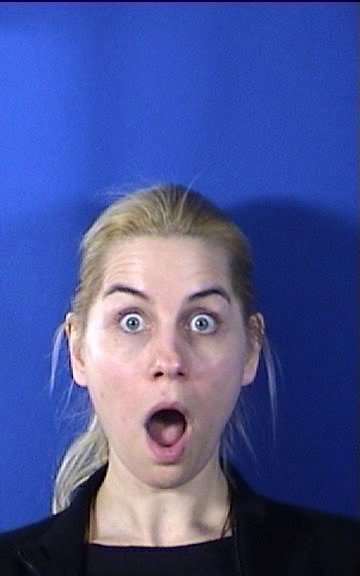}
      \caption*{Surprise}
    \end{subfigure}

\caption{This Figure shows the differences within the MMI dataset. The six used emotions are listed.}
\label{fig:mmi_images_dataset}
\end{figure}

\subsection{CKP Dataset}
\label{sec:ckp}

This dataset has been introduced by Lucey et al.~\cite{5543262}. 210 persons, aged 18 to 50, have been recorded depicting emotions.
This dataset presented by  contains recordings of emotions of 210 persons at the ages of 18 to 50 years. Both female and male persons are present and from different background. 81\% are Euro-Americans and 13\% are Afro-Americans. The images are of size 640$\times$490 px as well 640$\times$480 px. They are both grayscale and colored. In total this set has 593 emotion-labeled sequences. The emotions consist of \textit{Anger}, \textit{Disgust}, \textit{Fear}, \textit{Happiness}, \textit{Sadness}, \textit{Surprise}, and \textit{Contempt}.

\subsection{Comparison}
\label{sec:comparison}
In the MMI Dataset (Fig. \ref{fig:mmi_images_dataset}) the emotion \textit{Anger} is displayed in different ways, as can be seen by the eyebrows, forehead and mouth. The mouth in the lower image is tightly closed while in the upper image the mouth is open. For \textit{Disgust} the differences are also visible, as the woman in the upper picture has a much stronger reaction. The man depicting \textit{Fear} has contracted eyebrows which slightly cover the eyes. On the other hand the eyes of the woman are wide open. As for \textit{Happy} both persons are smiling strongly. In the lower image the woman depicting \textit{Sadness} has a stronger lip and chin reaction. The last emotion \textit{Surprise} also has differences like the openness of the mouth.\\

\begin{figure}
\centering

    \begin{subfigure}[b]{\mmiColWidth}
      \includegraphics[width=\textwidth]{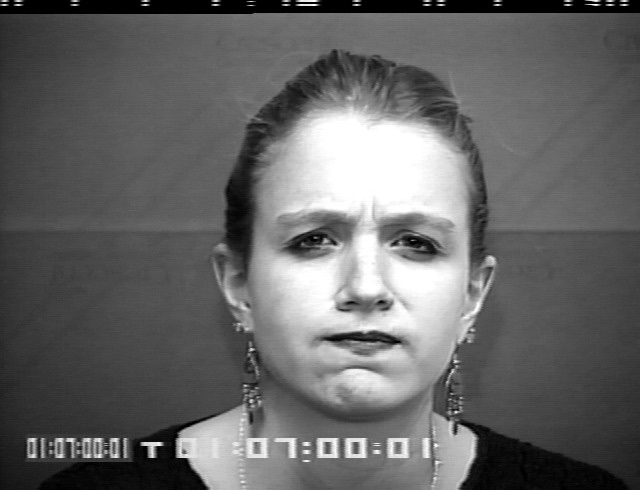}
    \end{subfigure}
    \begin{subfigure}[b]{\mmiColWidth}
      \includegraphics[width=\textwidth]{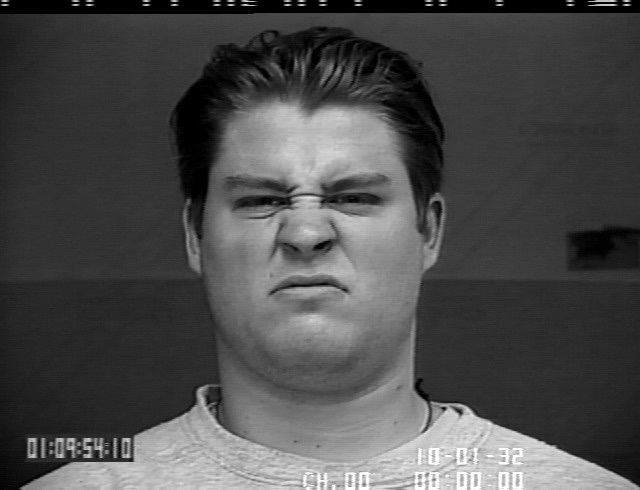}
    \end{subfigure}
    \begin{subfigure}[b]{\mmiColWidth}
      \includegraphics[width=\textwidth]{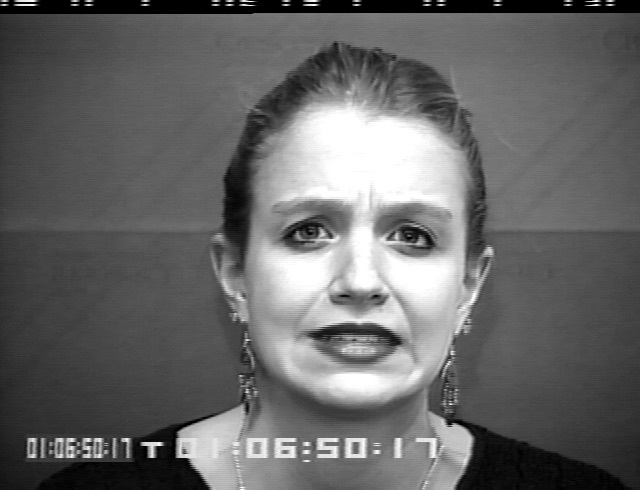}
    \end{subfigure}
    \begin{subfigure}[b]{\mmiColWidth}
      \includegraphics[width=\textwidth]{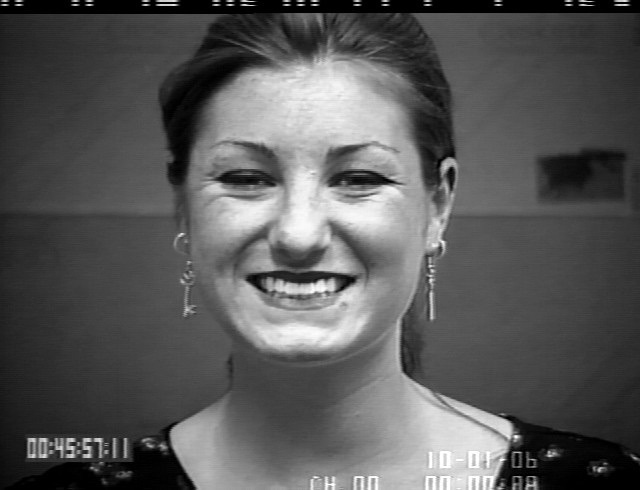}
    \end{subfigure}
    \begin{subfigure}[b]{\mmiColWidth}
      \includegraphics[width=\textwidth]{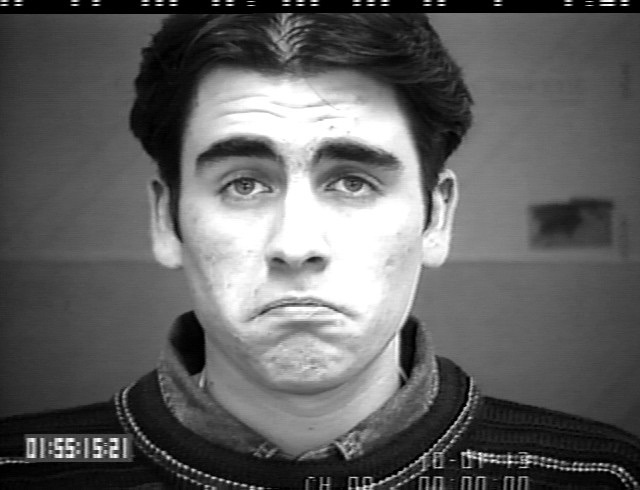}
    \end{subfigure}
    \begin{subfigure}[b]{\mmiColWidth}
      \includegraphics[width=\textwidth]{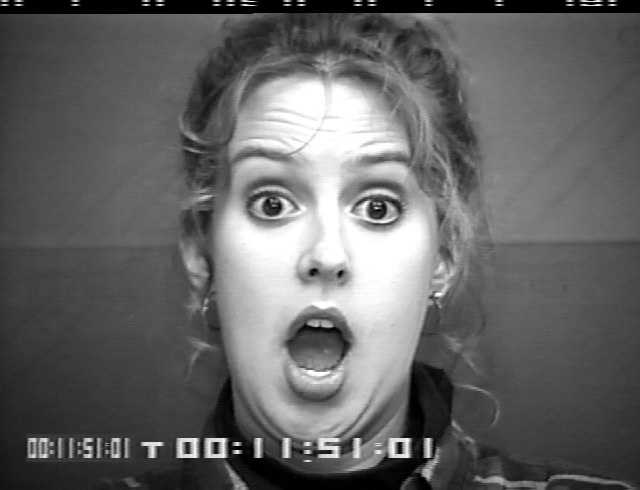}
    \end{subfigure}

    \begin{subfigure}[b]{\mmiColWidth}
      \includegraphics[width=\textwidth]{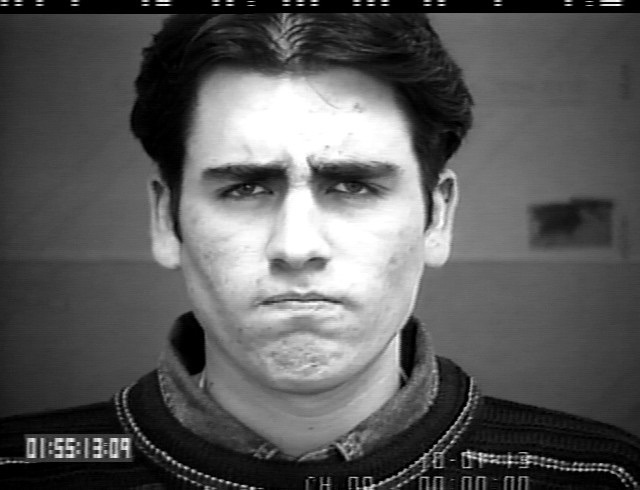}
      \caption*{\textit{Anger}}
    \end{subfigure}
    \begin{subfigure}[b]{\mmiColWidth}
      \includegraphics[width=\textwidth]{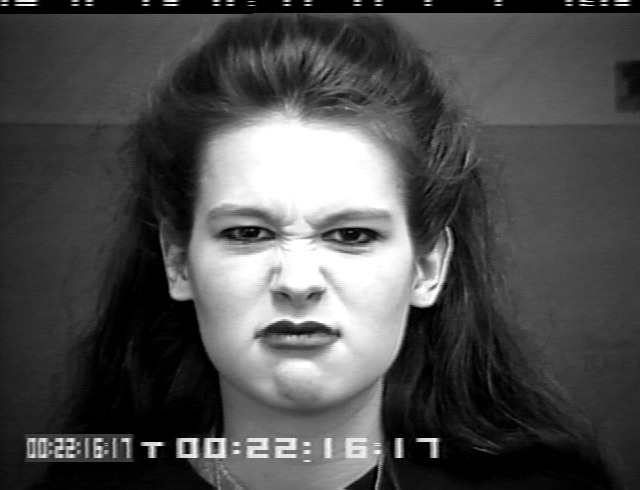}
      \caption*{\textit{Disgust}}
    \end{subfigure}
    \begin{subfigure}[b]{\mmiColWidth}
      \includegraphics[width=\textwidth]{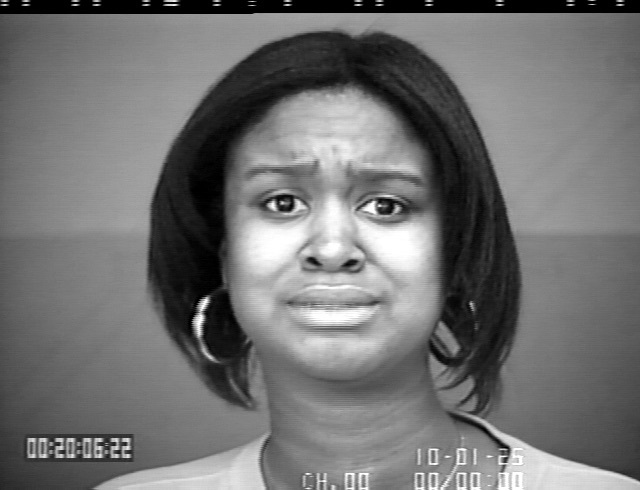}
      \caption*{\textit{Fear}}
    \end{subfigure}
    \begin{subfigure}[b]{\mmiColWidth}
      \includegraphics[width=\textwidth]{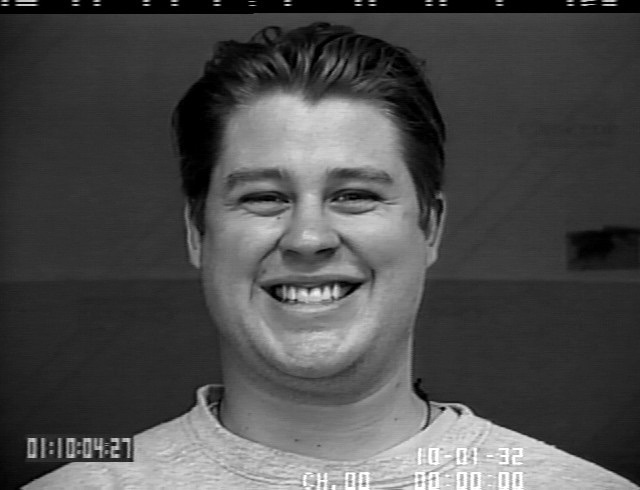}
      \caption*{\textit{Happy}}
    \end{subfigure}
    \begin{subfigure}[b]{\mmiColWidth}
      \includegraphics[width=\textwidth]{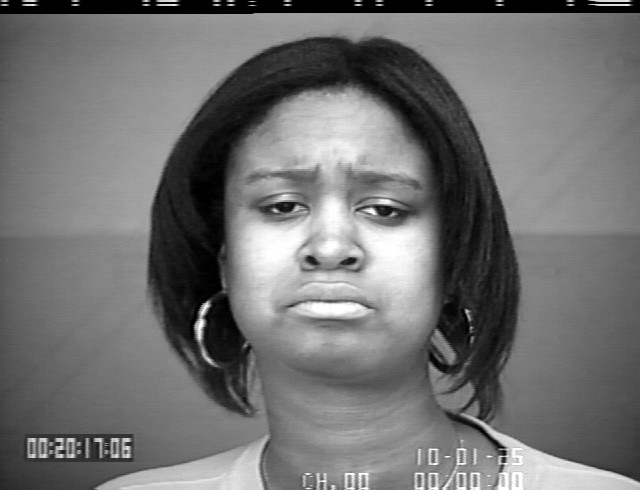}
      \caption*{\textit{Sadness}}
    \end{subfigure}
    \begin{subfigure}[b]{\mmiColWidth}
      \includegraphics[width=\textwidth]{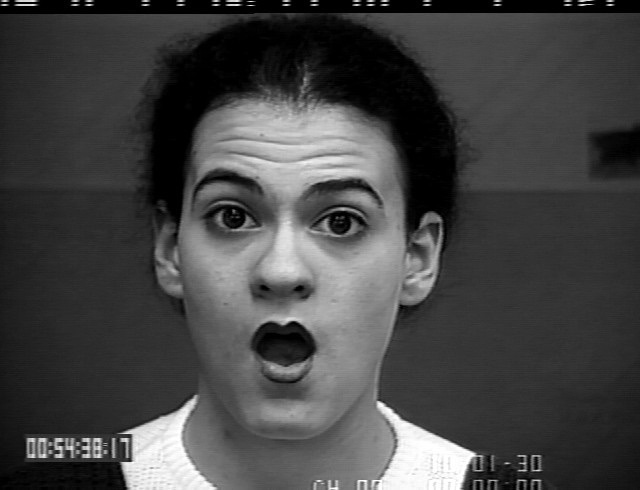}
      \caption*{\textit{Surprise}}
    \end{subfigure}

\caption{This Figure shows the differences within the Cohn-Kanade Plus (CKP) dataset. The emotion \textit{Contempt} is not shown since there is no annotated image with the emotion being depicted, which is allowed to be displayed.}
\label{fig:ckp_images_dataset}
\end{figure}

Such differences also appear in the CKP set (Fig. \ref{fig:ckp_images_dataset}). 
For \textit{Anger} the eyebrows and cheeks differ.
For \textit{Disgust} larger differences can be seen. In the upper picture not only the curvature of the mouth is stronger, but the nose is also more involved. While both women displaying \textit{Fear} show the same reaction around the eyes the mouth differs. In the lower image the mouth is nearly closed while teeth are visible in the upper one. \textit{Happiness} is displayed similar. For the emotion \textit{Sadness} the curvature of the mouth is visible in both images, but it is stronger in the upper one. The regions around the eyes differ as the eyebrows of the woman are straight. The last emotion \textit{Surprise} has strong similarities like the open mouth an wide open eyes. Teeth are only displayed by the woman in the upper image.\\
Thus for a better evaluation it is helpful to investigate multiple datasets. This aims at investigating whether the proposed approach works on different ways emotions are shown and whether it works on different emotions. For example \textit{Contempt} which is only included in the CKP set.

\begin{figure}[]
\centering
\includegraphics[scale=0.30]{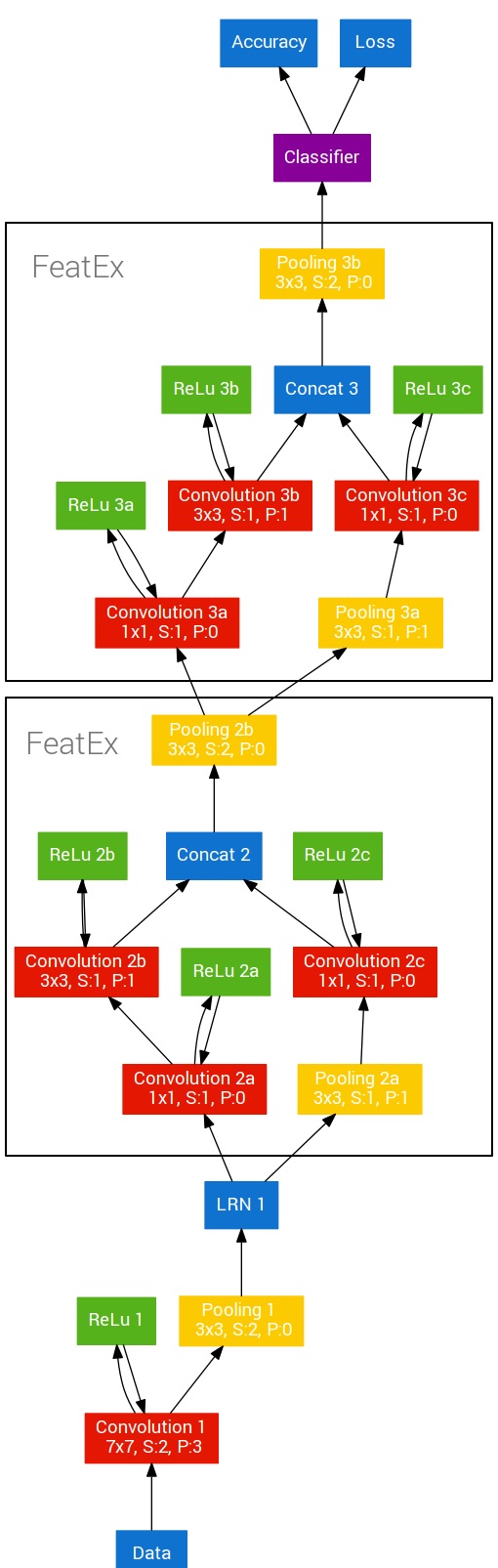}

\caption{This is the proposed architecture. The main component of this architecture is the FeatEx block. In the Convolutional layer $S$ depicts the Stride and $P$ the Padding.}
\label{fig:architecture}
\end{figure}

\section{Proposed Architecture}
\label{sec:architecture}

The proposed deep Convolutional Neural Network architecture (depicted in Figure~\ref{fig:architecture}) consists of four parts. The first part automatically preprocesses the data. This begins with Convolution 1, which applies $64$ different filters. The next layer is Pooling 1, which down-samples the images and then they are normalized by LRN 1. The next steps are the two FeatEx (Parallel Feature Extraction Block) blocks, highlighted in Figure~\ref{fig:architecture}. They are the core of the proposed architecture and described later in this section. The features extracted by theses blocks are forwarded to a fully connected layer, which uses them to classify the input into the different emotions.\\
The described architecture is compact, which makes it not only fast to train, but also suitable for real-time applications. This is also important as the network was built with resource usage in mind.

\begin{table}[ht!]
\centering
\caption{This Table lists the different output sizes produced by each layer.}
\label{tab:layers}

\begin{tabular}{l|l}
\hline
Layer & Output Size \\
\hline
Data & $224 \times 224$\\
Convolution 1 & $64 \times 112 \times 112$\\
Pooling 1 & $64 \times 56 \times 56$ \\
LRN 1 & $64 \times 56 \times 56$\\
\hline

Convolution 2a & $96 \times 56 \times 56$\\
Convolution 2b & $208 \times 56 \times 56$\\
Pooling 2a & $64 \times 56 \times 56$\\
Convolution 2c & $64 \times 56 \times 56$\\
Concat 2 & $272 \times 56 \times 56$\\
Pooling 2b & $272 \times 28 \times 28$\\
\hline

Convolution 3a & $96 \times 28 \times 28$\\
Convolution 3b & $208 \times 28 \times 28$\\
Pooling 3a & $272 \times 28 \times 28$\\
Convolution 3c & $64 \times 28 \times 28$\\
Concat 3 & $272 \times 28 \times 28$\\
Pooling 3b & $282 \times 14 \times 14$\\
\hline

Classifier & $11 \times 1 \times 1$\\
\hline
\end{tabular}
\end{table}

\paraV
\paragraph{\textit{FeatEx}}
The key structure in our architecture is the Parallel Feature Extraction Block (FeatEx). It is inspired by the success of GoogleNet. The block consists of Convolutional, Pooling, and ReLU Layers. The first Convolutional layer in FeatEx reduces the dimension since it convolves with a filter of size $1\times1$. It is enhanced by a ReLU layer, which creates the desired sparseness. The output is then convolved with a filter of size $3\times3$. In the parallel path a Max Pooling layer is used to reduce information before applying a CNN of size $1\times1$. This application of differently sized filters reflects the various scales at which faces can appear.\\
The paths are concatenated for a more diverse representation of the input. Using this block twice yields good results.\\

\begin{figure}[ht]
\centering
\includegraphics[width=\columnwidth]{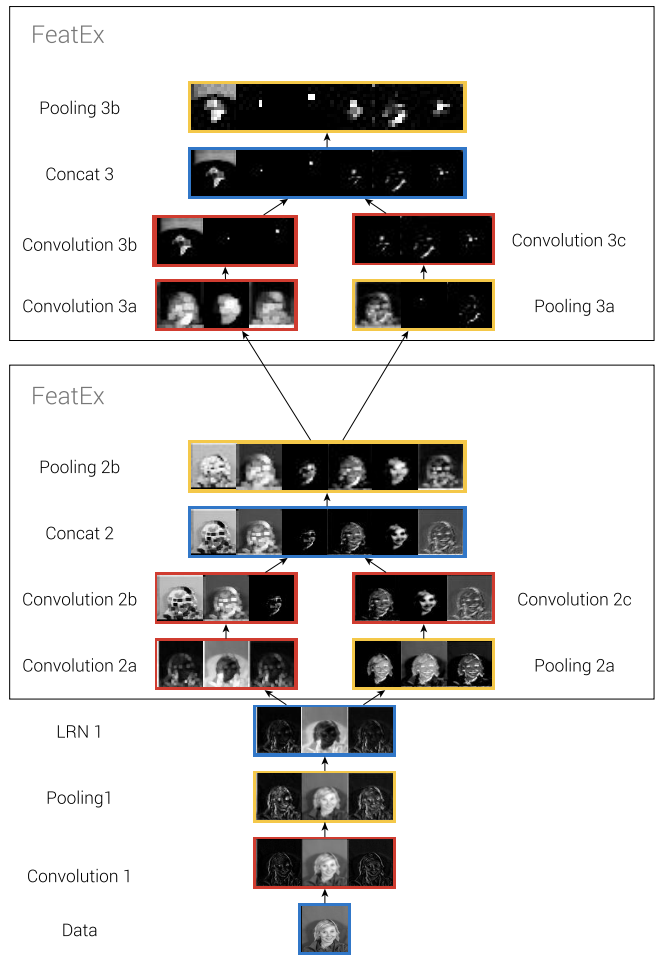}
\caption{This Figure shows example visualizations of the different layers. The data is taken from the MMI set.}
\label{fig:feature_extraction}
\end{figure}

\paraV
\paragraph{\textit{Visualization}}
The different layers of the architecture produce feature vectors as can be seen in Fig~\ref{fig:feature_extraction}.
The first part until LRN 1 preprocesses the data and creates multiple modified instances of the input. These show mostly edges with a low level of abstraction. The first FeatEx block creates two parallel paths of features with different scales, which are combined in Concat 2. The second FeatEx block refines the representation of the features. It also decreases the dimensionality.\\
This visualization shows that the concatenation of FeatEx blocks is a valid approach to create an abstract feature representation. The output dimensionality of each layer can be seen in Table \ref{tab:layers}.

\section{Experiments and Results}
\label{sec:experiments}

As implementation Caffe~\cite{jia2014caffe} was used. This is a deep learning framework, maintained by the Berkeley Vision and Learning Center (BVLC).

\begin{figure*}[ht!]
\centering
\begin{subfigure}[b]{0.46\textwidth}
\centering
\includegraphics[width=0.9\columnwidth]{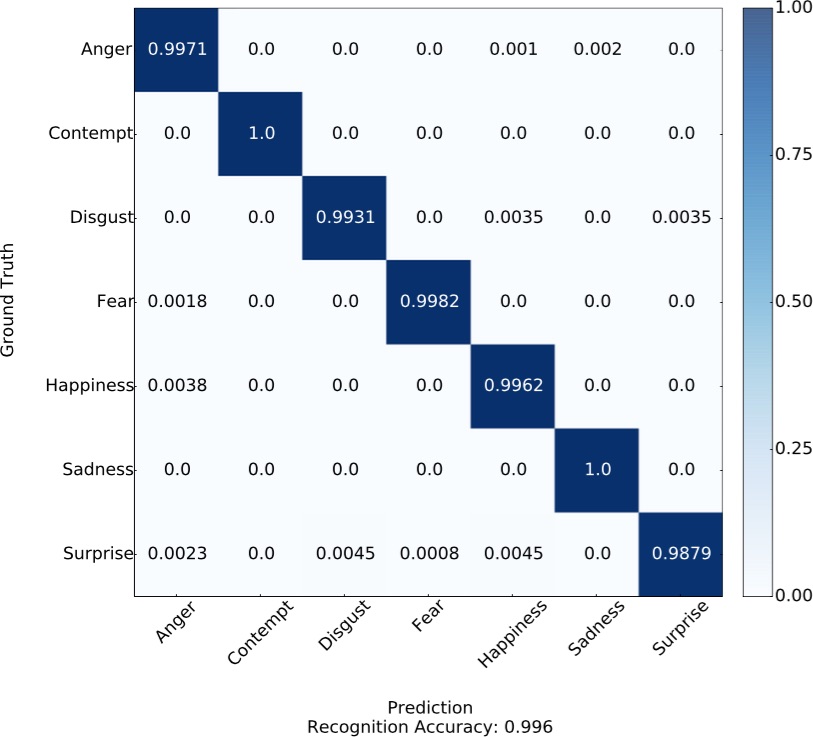}
\caption{The confusion matrix of the averaged 10-fold cross-validation on the CKP Dataset. The lowest accuracy is achieved by the emotion \textit{Surprise} with $98.79\%$ while \textit{Contempt}/\textit{Sadness} are both recognized with $100\%$.}
\label{fig:ckp_conf}
\end{subfigure}
\hspace{0.05\textwidth}
\begin{subfigure}[b]{0.46\textwidth}
\centering
\includegraphics[width=0.9\columnwidth]{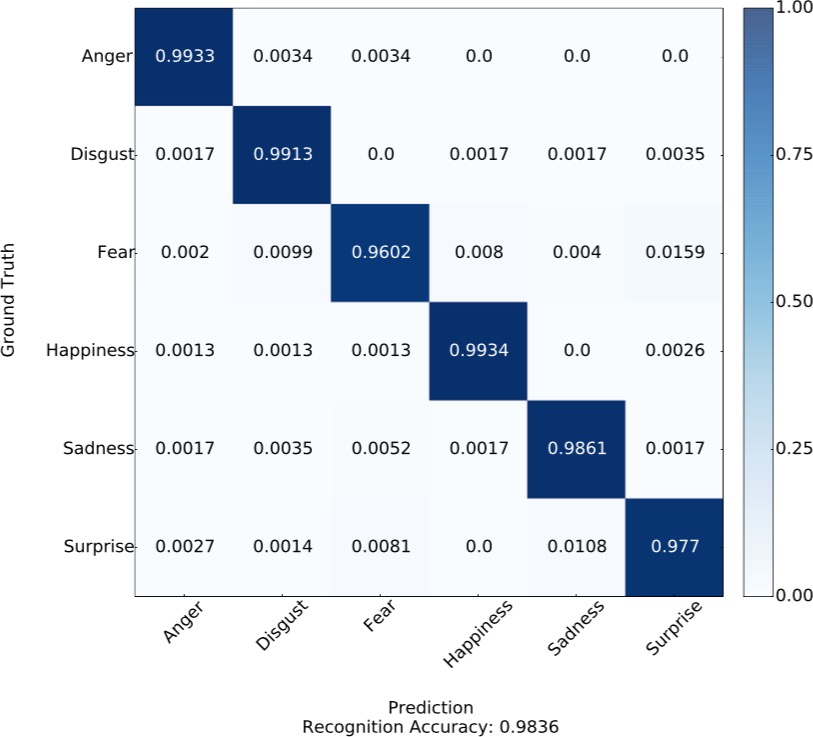}
\caption{The confusion matrix of the averaged 10-fold cross-validation on the MMI Dataset. The lowest accuracy is achieved by the emotion \textit{Fear} with $93.75\%$. \textit{Happiness} is recognized with $98.21\%$.\\~}
\label{fig:mmi_conf}
\end{subfigure}
\end{figure*}

\paraV
\paragraph{\textbf{CKP}}
The CKP database has been analyzed often and many different approaches have been evaluated in order to "solve" this set.
To determine whether the architecture is competitive, it has been evaluated on the CKP dataset. For the experiments all 5870 annotated images have been used to do a 10-fold cross-validation. The proposed architecture has proven to be very effective on this dataset with an average accuracy of 99.6\%. In Table~\ref{tab:ckp_results} different results from state of the art approaches are listed as comparison. The 100\% accuracy reported by Zafar~\cite{6743520} is based on hand picked images. The results are not validated using cross-validation. The confusion matrix in Fig.~\ref{fig:ckp_conf} depicts the results and shows that some emotions are perfectly recognized.

\begin{table}[b]
\centering
\caption{The CKP database has been very well analyzed and the best possible recognition accuracy has been achieved by Aliya Zafar. It is noteworthy that the samples he used for training are not randomly selected and no cross-validation has been applied. Evaluating this database provides information whether the proposed approach can compete with those results.}
\label{tab:ckp_results}
\resizebox{\columnwidth}{!}{
\begin{tabular}{l|c|c}
\hline
Author & Method & Accuracy \\
\hline
Aliya Zafar~\cite{6743520} & NCC & 100\%\\
Happy et al.~\cite{6998925} & Facial Patches + SVM & 94.09\%\\
Lucey et al.~\cite{5543262} & AAM + SVM & $\geq80\%$\\
Song et al.~\cite{6776135} & ANN (CNN) & 99.2\%\\
\hline
DeXpression(Proposed) & &99.6\% \\
\hline
\end{tabular}
}

\end{table}

\paraV
\paragraph{\textbf{MMI}}
The MMI Database contains videos of people showing emotions. From each video the 20 frames, which represent the content of the video the most, have been extracted fully automatically. The first two of these frames have been discarded since they provide neutral expressions. \\
To determine the frames, the difference between grayscale consecutive frames was calculated. To compensate noise the images have been smoothed using a Gaussian filter before calculation. To find the 20 most representative images, changes which occur in a small timeframe, should only be represented by a single image. This was achieved by iterating over the differences using a maximum filter with decreasing filter size until 20 frames have been found. In total 3740 images have been extracted.\\
The original images were then used for training and testing. A 10-fold cross-validation has been applied. The average accuracy is 98.63\%. This is better than the accuracies achieved by Wang and Yin~\cite{Wang:2007:STM:1287852.1288093} (Table~\ref{tab:mmi_results}). To our knowledge they have been the only ones to evaluate the MMI database on Emotions instead of Action Units. The results of the proposed approach are depicted in the Confusion Matrix in Fig.~\ref{fig:mmi_conf}. In the figure it is shown that the accuracy for \textit{Fear} is the lowest with 93.75\% while \textit{Happiness} is almost perfectly recognized with 98.21\%. \textit{Fear} and \textit{Surprise} are the emotions confused the most. 

\begin{table}[!ht]
\caption{This Table summarizes the current state of the art in emotion recognition on the MMI database (Section~\ref{sec:mmi}).}
\label{tab:mmi_results}
\centering
\begin{tabular}{l|c|c}
\hline
Author & Method & Accuracy \\
\hline
Wang and Yin~\cite{Wang:2007:STM:1287852.1288093} & LDA & 93.33\%\\
Wang and Yin~\cite{Wang:2007:STM:1287852.1288093} & QDC & 92.78\%\\
Wang and Yin~\cite{Wang:2007:STM:1287852.1288093} & NBC & 85.56\%\\
\hline
DeXpression (Proposed) & &98.63\% \\
\hline
\end{tabular}

\end{table}

\section{Discussion}
\label{sec:discussion}

The accuracy on the CKP set shows that the chosen approach is robust, misclassification usually occurs on pictures which are the first few instances of an emotion sequence. Often a neutral facial expression is depicted in those frames. Thus those misclassifications are not necessarily an error in the approach, but in the data selection. Other than that no major problem could be detected. The emotion \textit{Surprise} is often confused with \textit{Disgust} with a rate of 0.045\% which is the highest. Of those images, where an emotion is present, only few are wrongly classified.\\

As there is no consent for the misclassified images, they cannot be depicted here. However some unique names are provided. \\
Image S119\_001\_00000010 is classified as \textit{Fear} while the annotated emotion corresponds to \textit{Surprise}. The image depicts a person with a wide open mouth and open eyes. Pictures representing \textit{Surprise} are often very similar, since the persons also have wide open mouths and eyes. In image S032\_004\_00000014 the targeted label \textit{Fear} is confused with \textit{Anger}. While the mouth region in pictures with \textit{Anger} differ, the eye regions are alike, since in both situations the eyes and eyebrows are contracted.\\
Similar effects are experienced when dealing with the MMI Dataset. Since the first two frames are discarded most pictures with neutral positions are excluded. In few images a neutral position can still be found which gives rise to errors. For the same reason as the CKP set images will not be displayed. Due to the approach to extract images of the videos, a unique identifier for the misclassified image cannot be provided.\\
The top confusions are observed for \textit{Fear} and \textit{Surprise} with a rate of 0.0159\% where \textit{Fear} is wrongly misclassified as \textit{Surprise}. Session 1937 shows a woman displaying \textit{Fear} but it is classified as \textit{Surprise}. Both share common features like similar eye and mouth movement. In both emotions, participants move the head slightly backwards. This can be identified by wrinkled skin. The second most confusion rate, \textit{Surprise} being mistaken as \textit{Sadness}, is mostly based on neutral position images. Although the first two images are not used, some selected frames still do not contain an emotion. In Session 1985 \textit{Surprise} is being mistaken as \textit{Sadness}. The image depicts a man with his mouth being slightly curved, making him look sad.\\

DeXpression extracts features and uses them to classify images, but in very few cases the emotions are confused. This happens, as discussed, usually in pictures depicting no emotion. DeXpression performs very well on both tested sets, if an emotion is present.

\section{Conclusion and Future Work}
\label{sec:conclusion}

In this article DeXpression is presented which works fully automatically. It is a neural network which has little computational effort compared to current state of the art CNN architectures. In order to create it the new composed structure FeatEx has been introduced. It consists of several Convolutional layers of different sizes, as well as Max Pooling and ReLU layers. FeatEx creates a rich feature representation of the input. \\
The results of the 10-fold cross-validation yield, in average, a recognition accuracy of 99.6\% on the CKP dataset and 98.36\% on the MMI dataset. This shows that the proposed architecture is capable of competing with current state of the art approaches in the field of emotion recognition.\\
In Section~\ref{sec:discussion} the analysis has shown, that DeXpression works without major mistakes. Most misclassifications have occurred during the first few images of an emotion sequence. Often in these images emotions are not yet displayed.

\paraV
\paragraph{\textit{Future Work}}

An application built on DeXpression which is used in a real environment could benefit from distinguishing between more emotions such as \textit{Nervousness} and \textit{Panic}. Such a scenario could be large events where an early detection of \textit{Panic} could help to prevent mass panics. Other approaches to enhance emotion recognition could be to allow for composed emotions. For example frustration can be accompanied by anger, therefore not only showing one emotion, but also the reason. Thus complex emotions could be more valuable than basic ones. Besides distinguishing between different emotions, also the strength of an emotion could be considered. Being able to distinguish between different levels could improve applications, like evaluating reactions to new products. In this example it could predict the amount of orders that will be made, therefore enabling producing the right amount of products.

\section*{Acknowledgments}

We would like to thank the Affect Analysis Group of the University of Pittsburgh for providing the Extended Cohn–Kanade database, and Prof. Pantic and Dr. Valstar for the MMI data-base.

\bibliographystyle{IEEEtran}
\bibliography{list}

\begin{thebibliography}{10}
\providecommand{\url}[1]{#1}
\csname url@samestyle\endcsname
\providecommand{\newblock}{\relax}
\providecommand{\bibinfo}[2]{#2}
\providecommand{\BIBentrySTDinterwordspacing}{\spaceskip=0pt\relax}
\providecommand{\BIBentryALTinterwordstretchfactor}{4}
\providecommand{\BIBentryALTinterwordspacing}{\spaceskip=\fontdimen2\font plus
\BIBentryALTinterwordstretchfactor\fontdimen3\font minus
  \fontdimen4\font\relax}
\providecommand{\BIBforeignlanguage}[2]{{%
\expandafter\ifx\csname l@#1\endcsname\relax
\typeout{** WARNING: IEEEtran.bst: No hyphenation pattern has been}%
\typeout{** loaded for the language `#1'. Using the pattern for}%
\typeout{** the default language instead.}%
\else
\language=\csname l@#1\endcsname
\fi
#2}}
\providecommand{\BIBdecl}{\relax}
\BIBdecl

\bibitem{Anderson06areal-time}
K.~Anderson and P.~W. Mcowan, ``A real-time automated system for recognition of
  human facial expressions,'' \emph{IEEE Trans. Syst., Man, Cybern. B, Cybern},
  pp. 96--105, 2006.

\bibitem{4032815}
I.~Kotsia and I.~Pitas, ``Facial expression recognition in image sequences
  using geometric deformation features and support vector machines,''
  \emph{Image Processing, IEEE Transactions on}, vol.~16, no.~1, pp. 172--187,
  Jan 2007.

\bibitem{kumar2009face}
B.~V. Kumar, ``Face expression recognition and analysis: the state of the
  art,'' \emph{Course Paper, Visual Interfaces to Computer}, 2009.

\bibitem{6998925}
S.~Happy and A.~Routray, ``Automatic facial expression recognition using
  features of salient facial patches,'' \emph{Affective Computing, IEEE
  Transactions on}, vol.~6, no.~1, pp. 1--12, Jan 2015.

\bibitem{donahue2013decaf}
J.~Donahue, Y.~Jia, O.~Vinyals, J.~Hoffman, N.~Zhang, E.~Tzeng, and T.~Darrell,
  ``Decaf: A deep convolutional activation feature for generic visual
  recognition,'' \emph{arXiv preprint arXiv:1310.1531}, 2013.

\bibitem{DBLP:journals/corr/SzegedyLJSRAEVR14}
\BIBentryALTinterwordspacing
C.~Szegedy, W.~Liu, Y.~Jia, P.~Sermanet, S.~Reed, D.~Anguelov, D.~Erhan,
  V.~Vanhoucke, and A.~Rabinovich, ``Going deeper with convolutions,''
  \emph{CoRR}, vol. abs/1409.4842, 2014. [Online]. Available:
  \url{http://arxiv.org/abs/1409.4842}
\BIBentrySTDinterwordspacing

\bibitem{LSVRC-results}
\BIBentryALTinterwordspacing
LSVRC. (2014, Jun.) Results of the lsvrc challenge. [Online]. Available:
  \url{http://www.image-net.org/challenges/LSVRC/2014/results}
\BIBentrySTDinterwordspacing

\bibitem{caleanu2013face}
C.-D. Caleanu, ``Face expression recognition: A brief overview of the last
  decade,'' in \emph{Applied Computational Intelligence and Informatics (SACI),
  2013 IEEE 8th International Symposium on}.\hskip 1em plus 0.5em minus
  0.4em\relax IEEE, 2013, pp. 157--161.

\bibitem{bettadapura2012face}
V.~Bettadapura, ``Face expression recognition and analysis: the state of the
  art,'' \emph{arXiv preprint arXiv:1203.6722}, 2012.

\bibitem{byeonfacial}
Y.-H. Byeon and K.-C. Kwak, ``Facial expression recognition using 3d
  convolutional neural network.''

\bibitem{song2014deep}
I.~Song, H.-J. Kim, and P.~B. Jeon, ``Deep learning for real-time robust facial
  expression recognition on a smartphone,'' in \emph{Consumer Electronics
  (ICCE), 2014 IEEE International Conference on}.\hskip 1em plus 0.5em minus
  0.4em\relax IEEE, 2014, pp. 564--567.

\bibitem{5543262}
P.~Lucey, J.~Cohn, T.~Kanade, J.~Saragih, Z.~Ambadar, and I.~Matthews, ``The
  extended cohn-kanade dataset (ck+): A complete dataset for action unit and
  emotion-specified expression,'' in \emph{Computer Vision and Pattern
  Recognition Workshops (CVPRW), 2010 IEEE Computer Society Conference on},
  June 2010, pp. 94--101.

\bibitem{Shan2009803}
C.~Shan, S.~Gong, and P.~W. McOwan, ``Facial expression recognition based on
  local binary patterns: A comprehensive study,'' \emph{Image and Vision
  Computing}, vol.~27, no.~6, pp. 803 -- 816, 2009.

\bibitem{6743520}
A.~Zafer, R.~Nawaz, and J.~Iqbal, ``Face recognition with expression variation
  via robust ncc,'' in \emph{Emerging Technologies (ICET), 2013 IEEE 9th
  International Conference on}, Dec 2013, pp. 1--5.

\bibitem{AISTATS2011_GlorotBB11}
X.~Glorot, A.~Bordes, and Y.~Bengio, ``Deep sparse rectifier neural networks,''
  in \emph{Proceedings of the Fourteenth International Conference on Artificial
  Intelligence and Statistics (AISTATS-11)}, G.~J. Gordon and D.~B. Dunson,
  Eds., vol.~15.\hskip 1em plus 0.5em minus 0.4em\relax Journal of Machine
  Learning Research - Workshop and Conference Proceedings, 2011, pp. 315--323.

\bibitem{Pantic2005wdffe}
M.~Pantic, M.~F. Valstar, R.~Rademaker, and L.~Maat, ``Web-based database for
  facial expression analysis,'' in \emph{Proceedings of IEEE Int'l Conf.
  Multimedia and Expo (ICME'05)}, Amsterdam, The Netherlands, July 2005, pp.
  317--321.

\bibitem{jia2014caffe}
Y.~Jia, E.~Shelhamer, J.~Donahue, S.~Karayev, J.~Long, R.~Girshick,
  S.~Guadarrama, and T.~Darrell, ``Caffe: Convolutional architecture for fast
  feature embedding,'' \emph{arXiv preprint arXiv:1408.5093}, 2014.

\bibitem{6776135}
I.~Song, H.-J. Kim, and P.~Jeon, ``Deep learning for real-time robust facial
  expression recognition on a smartphone,'' in \emph{Consumer Electronics
  (ICCE), 2014 IEEE International Conference on}, Jan 2014, pp. 564--567.

\bibitem{Wang:2007:STM:1287852.1288093}
J.~Wang and L.~Yin, ``Static topographic modeling for facial expression
  recognition and analysis,'' \emph{Comput. Vis. Image Underst.}, vol. 108, no.
  1-2, pp. 19--34, Oct. 2007.

\end{thebibliography}

\end{document}